\documentclass[letterpaper, 10 pt, conference]{ieeeconf}

\IEEEoverridecommandlockouts
\usepackage[english]{babel}
\usepackage{amsmath}  
\usepackage{amssymb}  
\usepackage{graphicx}
\usepackage{subfigure}
\usepackage{caption}
\usepackage{multirow}
\usepackage{multicol}
\usepackage{amsfonts}
\usepackage{tabularx}
\usepackage{algorithm,algorithmic}
\usepackage{bm}
\usepackage{breakurl}
\usepackage{enumerate}
\usepackage{adjustbox}
\usepackage{xcolor}
\usepackage{siunitx}
\newcolumntype{b}{X}
\newcolumntype{s}{>{\hsize=.3\hsize}X}
\newcolumntype{x}{>{\hsize=.5\hsize}X}
\usepackage{tikz}
\usetikzlibrary{shapes.geometric, arrows}
\usepackage{array}
\usepackage{cite}
\usepackage{wasysym}
\usepackage{url}
\usepackage{xurl}
\usepackage{booktabs}
\makeatletter
\let\NAT@parse\undefined
\makeatother
\usepackage{hyperref}
\usepackage{cleveref}

\tikzstyle{startstop} = [rectangle, rounded corners, minimum width=3cm, minimum height=1cm,text centered, draw=black, fill=red!30]
\tikzstyle{arrow} = [thick,->,>=stealth]

\title{\LARGE \bf Experimental System Design of an Active Fault-Tolerant Quadrotor}

\author{Jennifer Yeom, Roshan Balu T M B, Guanrui Li, and Giuseppe Loianno
\thanks{The authors are with the New York University, Tandon School of Engineering, Brooklyn, NY 11201, USA. {\tt\footnotesize email: \{jennifer.yeom, rt2420, gl1871, loiannog\}@nyu.edu}.}
\thanks{This work was supported by the NSF CAREER Award 2145277, the DARPA YFA Grant D22AP00156-00, the NSF CPS Grant CNS-2121391, Qualcomm Research, Nokia, and NYU Wireless.}
}

\begin{document}
\makeatletter
\g@addto@macro\@maketitle{
\setcounter{figure}{0}
\centering
    \includegraphics[width=0.99\textwidth]{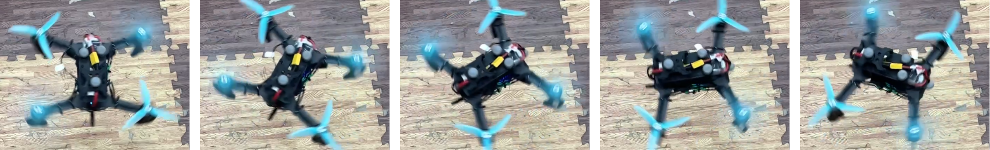}
	\captionof{figure}{A top-down view of our quadrotor designed specifically for fault-tolerant control in rapid rotation (clockwise) as 2 of its motors are subject to failure. This quadrotor is spinning at a rate of 25~radians/sec (equivalent to 1430~degrees/sec or 4 revolutions/sec). \label{fig:intro}}
}
\makeatother

\maketitle
\thispagestyle{empty}
\pagestyle{empty}

\begin{abstract}
Quadrotors have gained popularity over the last decade, aiding humans in complex tasks such as search and rescue, mapping and exploration. 
Despite their mechanical simplicity and versatility compared to other types of aerial vehicles, they remain vulnerable to rotor failures. 
In this paper, we propose an algorithmic and mechanical approach to addressing the quadrotor fault-tolerant problem in case of rotor failures. First, we present a fault-tolerant detection and control scheme that includes various attitude error metrics. The scheme transitions to a fault-tolerant control mode by surrendering the yaw control. Subsequently, to ensure compatibility with platform sensing constraints, we investigate the relationship between variations in robot rotational drag, achieved through a modular mechanical design appendage, resulting in yaw rates within sensor limits. This analysis offers a platform-agnostic framework for designing more reliable and robust quadrotors in the event of rotor failures.
Extensive experimental results validate the proposed approach providing insights into successfully designing a cost-effective quadrotor capable of fault-tolerant control. The overall design enhances safety in scenarios of faulty rotors, without the need for additional sensors or computational resources.


\end{abstract}


\IEEEpeerreviewmaketitle

\section{Introduction} \label{sec:introduction}
Quadrotors have emerged as widely-used robots in various applications ranging from disaster response~\cite{almurib2011searchandrescue} to transportation~\cite{li2021transportation}. 
However, their potential utility is accompanied by inherent challenges, notably their underactuated nature. 
While underactuation empowers quadrotors with mechanical simplicity, it also introduces potential of instability and control challenges, especially when one or more rotors fail. 
This vulnerability underscores the importance of innovative strategies to safely and effectively harness their capabilities.

In our previous works~\cite{yeom2023geometric,mao2023propeller}, we contributed to an active fault-tolerant control method for a quadrotor, which identifies the system faults and modifies the control strategy accordingly to compensate for the rotor failures~\cite{abbaspour2020survey}. 
Moreover, in~\cite{yeom2023geometric}, we investigated four distinct attitude error metrics to evaluate the feasibility of fault-tolerant control in scenarios involving one or two failed rotors.
Through simulation, we showed in~\cite{yeom2023geometric} that error metrics that exclude yaw information, known as reduced attitude metrics~\cite{chaturvedi2011attitudecontrol}, performed better compared to those incorporating yaw information. 
However, these metrics were not evaluated in real-world experiments in~\cite{yeom2023geometric}, which is vital for understanding their impact on a quadrotor's fault-tolerant control under rotor failures. 
Additionally, our previous research~\cite{mao2023propeller} revealed that combining an appropriate mechanical design and control structure could potentially facilitate robust fault-tolerant control as well as the transition to fault-tolerant control in rotor failure scenarios, yet no detailed analysis was provided.

Therefore, this paper presents several key contributions. 
First, we introduce and validate a fault-tolerant control scheme that combines multiple attitude error metrics with a fault-tolerant detection module inspired by~\cite{mao2023propeller}. Second, drawing from our empirical observations during real-world experiments, we study the intricate connection between the introduction of augmented drag obtained through a modular mechanical design appendage and the resulting yaw rate of the compromised quadrotor, aiming to enhance steady-state angular velocity dynamics for flight safety. Specifically, we analyze the relationship between the drag coefficient estimation and the area of the appendage with the goal to offer a platform-agnostic guideline for designing safer and more robust quadrotors in the event of rotor failures.
Finally, we provide extensive experimental results that validate both the proposed control scheme and drag mechanical design. We study the flight envelope under the proposed fault-tolerant control scheme and identify its limits for high speeds, roll and pitch angles. We also experimentally identify the effects of latency on the effectiveness of the transition and corresponding fault-tolerant control modality. Furthermore, the results on the augmented drag and its drag estimation offer deep insights into quadrotor mechanical design in the event of rotor failures, enhancing the resilience of the proposed fault-tolerant control scheme. 

The rest of this paper is organized as follows. Section~\ref{sec:related_works} introduces the state of the art in the area of fault-tolerant control and fault detection for aerial robots specifically focusing on rotor failures. 
Section~\ref{sec:methodology} describes our active fault-tolerant control approach and discusses the choice of different attitude metrics along with critical design considerations. 
Additionally, we detail our methodology for estimating drag and highlight key design considerations.
Section~\ref{sec:results} shows the experimental results and detailed numerical and comparative analysis. 
Section~\ref{sec:conclusion} concludes the paper and presents potential directions for future works in this area. 

\section{Related Works}~\label{sec:related_works}
Fault tolerance has remained an enduring challenge at the forefront of both the robotics and aerospace industries.
In this section, we address two major challenges to active fault-tolerant control: fault-tolerant control strategies and fault detection methods. 

\subsection{Fault-Tolerant Control}
Most of the fault-tolerant control strategies for quadrotors surrender the control in yaw and yaw rate when dealing with a failed rotor or propeller. 
Many works show the theoretical possibility of fault-tolerant control only in simulations~\cite{lippiello2014emergency,freddi2011feedback,sharifi2010slidingmode,deCrousaz2015sql}, where aspects such as thrust-to-weight-ratio, actuator limits, or angular rate of a rigid body do not pose physical constraints.
These works include but are not limited to proportional-integral-derivative (PID) control~\cite{lippiello2014emergency}, feedback linearization~\cite{freddi2011feedback}, sliding mode~\cite{sharifi2010slidingmode}, and Sequential Linear Qaudratic (SLQ) control~\cite{deCrousaz2015sql}.

Several works have conducted real-world flight tests involving failed rotors or propellers. 
One of the earliest works~\cite{mueller2014lqr} uses a Linear Quadratic Regulator (LQR) to solve for periodic solutions around a linearized point for failed rotor scenarios. 
An Incremental Nonlinear Dynamic Inversion (INDI) approach is used in~\cite{sun2021indi} showing the successful position control of a quadrotor with the loss of two opposing rotors in the presence of significant wind disturbances. 
The fault-tolerant controller in~\cite{nan2022npmcfault} uses Model Predictive Control (MPC) to recover a quadrotor from highly dynamic and inverted positions with a rotor failure.
These works focus on the control of a quadrotor with rotor failure(s); however, do not consider the detection of the fault, potential latency issues from a fault to detection, nor the transition dynamics due to latency.
Our controller design considers not only the detection and transition to fault-tolerant control but also addresses the intricacies of quadrotor dynamics during this transition.
We also study the rotational drag of the system to provide platform-agnostic design guidelines for quadrotors capable of flight with one or two rotor failures.

\subsection{Fault Detection}
Many existing fault detection methods for actuator faults involve estimation techniques~\cite{Madruga2023LOEAero, Avram_2017_detect_adapt, schijndel2023eject}. 
A reduced order Extended Kalman Filter (EKF) is used in~\cite{Madruga2023LOEAero} to estimate the Loss of Effectiveness (LOE) of an actuator in the presence of aerodynamic effects. 
Real experiments show accurate estimation when the quadrotor experiences up to $15\%$ LOE, emulated by slowing down a motor by $85\%$ of its original speed.
Nonlinear adaptive estimators are used in~\cite{Avram_2017_detect_adapt} to detect, diagnose, and adapt to quadrotor actuator failures up to $10\%$ in magnitude. 
A propeller ejection mechanism is developed to mimic a catastrophic failure of a rotor in ~\cite{schijndel2023eject}. 
A Kalman filter is used to detect the complete loss of a propeller while using the RPM (revolutions per minute) readings from the motor as an input. 

Other works use empirical methods to estimate and detect a system fault.
The vibration noise of an accelerometer is used to estimate propeller damage in~\cite{vibrations_diagonosis}, but is only tested with very small amounts of damage.
Parameter identification techniques are used in~\cite{luLOE} to accurately estimate the LOE of a rotor in simulations.
Although these estimation and empirical methods showcase successful fault detection and diagnosis, none of them show the ability to transition to a fault-tolerant controller when necessary. 
Rather, they only compensate to small amounts (up to $15\%$ loss of effectiveness) of actuator or propeller failures. 

Limited works exist that handle both fault detection and fault-tolerant control, also known as active fault-tolerant control.
The LOE of a motor is used for fault detection and transition to a fault-tolerant controller in~\cite{luLOE}. 
However, only simulation results are shown in this work.
Overall, there is a distinct lack in literature of research that integrates fault detection and an autonomous transition to fault-tolerant control.
Our work shows a fully autonomous fault detection and transition from the standard to the fault-tolerant controller.
We also experimentally demonstrate the required reaction time for a fault detection and transition module to effectively recover a faulty quadrotor.
This is a crucial element in the seamless integration of fault detection and control methods.
\begin{figure}
    \centering
    \vspace{0.2cm}
    \includegraphics[width=\linewidth]{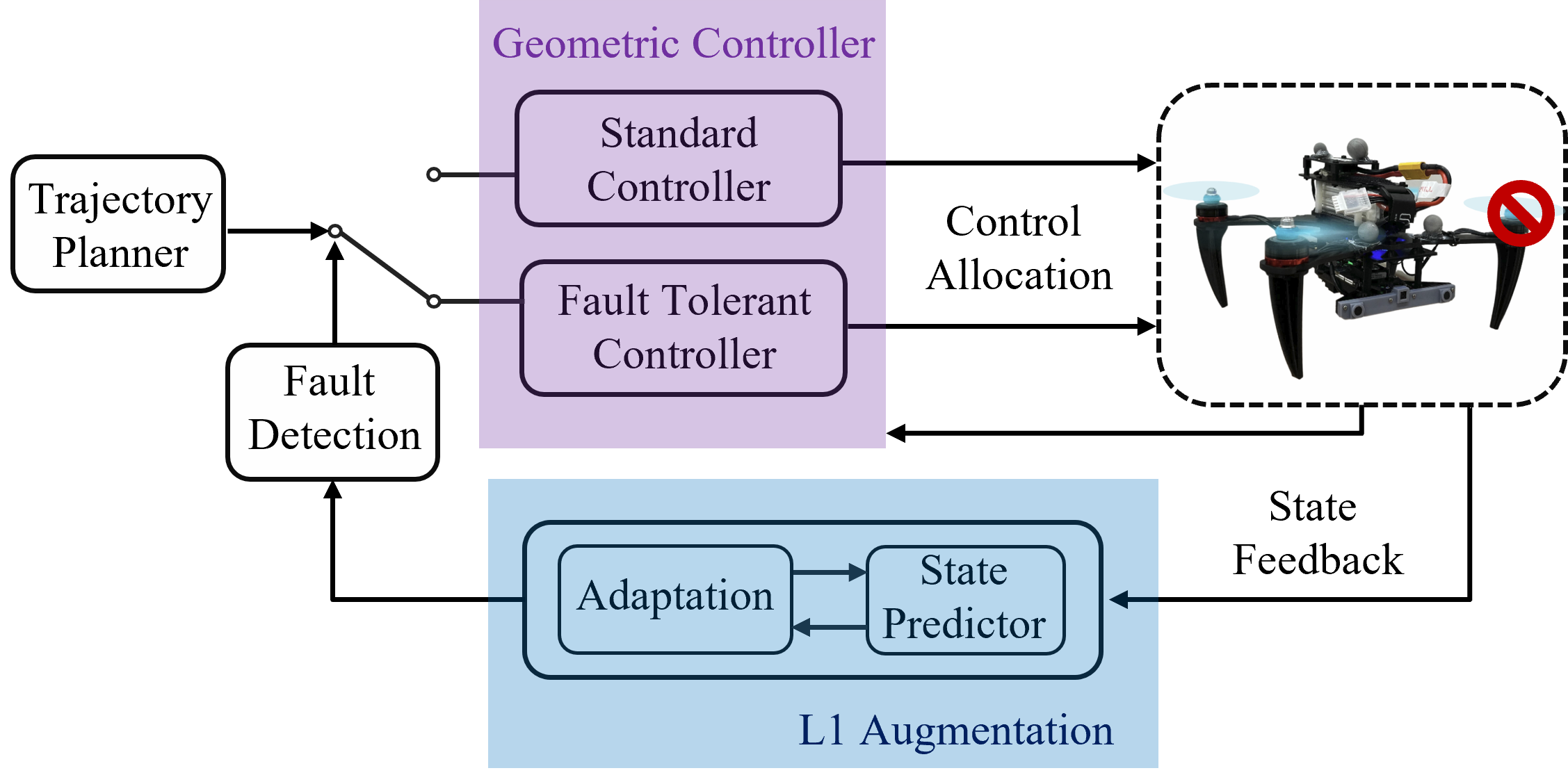}
    \caption{Overview of our active fault-tolerant control scheme with state prediction, adaptive, and fault detection modules. The L1 augmentation calculates a damage estimate and is passed to the fault detection threshold which dictates the switch between a standard and fault-tolerant controller.}
    \vspace{-0.3cm}
    \label{fig:control_diag}
\end{figure}

\section{Methodology} \label{sec:methodology}
In this section, we introduce our active fault-tolerant control scheme and design considerations of a fault-tolerant control capable quadrotor.
First, we implement and compare three distinct reduced attitude error metrics, omitting yaw control in favor of a fault-tolerant solution. 
Next, we employ a fault detection algorithm, utilizing a state predictor and an adaptive policy~\cite{mao2023propeller}, enabling an automated transition from a normal controller to a fault-tolerant counterpart. 
We also investigate the impact of drag on the quadrotor and its influence on the steady-state yaw rate. 
The estimation strategy of the resulting drag coefficient from the added drag is also presented. 
An overview block diagram of the control architecture is shown in Fig.~\ref{fig:control_diag}.

\subsection{Preliminaries}
\begin{figure}
    \centering
    \includegraphics[width=0.8\linewidth]{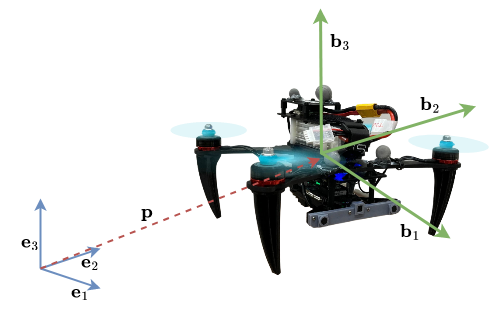}
    \caption{Quadrotor model with inertial (blue) and body (green) frame definitions. Both frames are right-hand-frames.}
    \label{fig:quadrotor_model}
    \vspace{-0.3cm}
\end{figure}

The quadrotor used in our real-world experiments is shown in Fig.~\ref{fig:quadrotor_model} along with the definitions of the inertial and body frames. 
The equations of motion of a quadrotor can be written as
\begin{align} \label{EoM}
\begin{split}
    \dot{\mathbf{p}} &= \mathbf{v},\\
    f \mathbf{R} \mathbf{e}_3 &= m \dot{\mathbf{v}} + m g \mathbf{e}_3 \\
    \dot{\mathbf{R}} &= \mathbf{R} \hat{\mathbf{\Omega}},\\
    \mathbf{M} &= \mathbf{J} \dot{\mathbf{\Omega}} + \mathbf{\Omega} \times \mathbf{J}\mathbf{\Omega} + \mathbf{D},
\end{split}
\end{align}
where $\mathbf{p} = \begin{bmatrix}x & y & z \end{bmatrix}^\top$, $\mathbf{v} = \begin{bmatrix}\dot{x}& \dot{y}& \dot{z}\end{bmatrix}^\top$ are the position and velocity of the quadrotor.
The variables $m$, $g$, $f$ are the mass, gravity, and thrust.
$\mathbf{R}$, $\mathbf{\Omega} = \begin{bmatrix} \Omega_1 & \Omega_2 & \Omega_3 \end{bmatrix}^\top$, $\mathbf{M}=\begin{bmatrix} M_1 & M_2 & M_3 \end{bmatrix} ^\top$, $\mathbf{J}$, $\mathbf{D}$ are the orientation, angular velocity, moment, inertia of the quadrotor, and the rotational drag term respectively.
Specifically, the third term of $\mathbf{D}$, which we will note as $\tau_{drag}$ and how it relates to a the fault-tolerant dynamics will be explored further in our methodology. 
The \textit{hat map}, $\hat{\cdot}$ : $\mathbb{R}^{3} \rightarrow so(3)$  represents the mapping such that $\hat{\mathbf{a}}\mathbf{b} = \mathbf{a} \times \mathbf{b}, \forall \ \mathbf{a},\mathbf{b} \in \mathbb{R}^{3}$.

\subsection{Fault-Tolerant Control} \label{sec:control}

We choose to use reduced attitude error metrics in our implementation of fault-tolerant control as we surrender the control of yaw.
Consider $\mathbf{n} = \mathbf{R}^\top \mathbf{d}$ where $\mathbf{d}$ is a unit vector and $\mathbf{d} \in \mathcal{S}^{2}$. 
This representation states that the inertial unit vector $\mathbf{n}$ is obtained by applying the transpose of the rotation matrix $\mathbf{R}$ to the body-fixed unit vector $\mathbf{d}$.  
This matrix multiplication accounts for the rotation of the body-fixed vector into the inertial frame. 
The rotation matrix $\mathbf{R} = \begin{bmatrix} \mathbf{b}_1& \mathbf{b}_2& \mathbf{b}_3 \end{bmatrix} \in SO(3)$, in this case, is determined by the axis and angle of rotation, and can be decomposed by
\begin{equation} \label{eq:decomposition}
\mathbf{R} = \mathbf{R}_{\psi} \mathbf{R}_{\phi}, 
\end{equation}
\begin{equation} \label{eq:R_yaw}
\mathbf{R}_{\psi} = 
\begin{bmatrix}
 \cos(\psi)& -\sin(\psi) & 0\\ 
 \sin(\psi)& \cos(\psi) & 0 \\ 
 0& 0 & 1
\end{bmatrix},
\end{equation}
\begin{equation} \label{eq:R_tilt}
\mathbf{R}_{\phi} = 
\begin{bmatrix}
b_{3z} + \frac{b_{3y}^2}{1+b_{3z}} & \frac{-b_{3x} b_{3y}}{1 + b_{3z}} & b_{3x}\\ 
\frac{-b_{3x} b_{3y}}{1 + b_{3z}}  & 1 - \frac{b_{3y}}{1+b_{3z}}  & b_{3y}\\ 
-b_{3x} & -b_{3y} & b_{3z}
\end{bmatrix}.
\end{equation}
The tilt ($\mathbf{R}_{\phi}$) is the direction of the unit vector about which the rotation occurs, and the angle of rotation ($\mathbf{R}_{\psi}$) is the magnitude of the rotation about that axis. 
In our purposes of a fault-tolerant controller, we utilize this reduced attitude to control the quadrotor's attitude without considering its yaw state. 
In other words, the fault-tolerant controller only considers where the rigid body is pointing, not the rotation about the pointing direction.

We alter the geometric controller introduced in~\cite{lee2010so3} to use only a reduced attitude for our purpose in a fault-tolerant controller. 
For brevity, we only show the adjustment made in the attitude error $\mathbf{e}_R$ term when solving for moments in 
\begin{equation}
\mathbf{M} = -\mathbf{k}_R \mathbf{e}_R - \mathbf{k}_{\Omega} \mathbf{e}_{\Omega} + \mathbf{\Omega}\times \mathbf{J} \mathbf{\Omega}.
\label{eq:moment_commands}
\end{equation}
Additional details of the fault-tolerant controller can be found in~\cite{yeom2023geometric}. Next, we describe the three different reduced attitude metrics applied and evaluated: Current Yaw Metric~($\mathbf{e}_{cy}$), $\mathcal{S}^{2}$ Metric~($\mathbf{e}_{S^2}$), and Thrust Vector Metric~($\mathbf{e}_{tv}$).

\subsubsection{Current Yaw Metric} 
For our first reduced attitude error metric, we simply use the skew symmetric definition of the geometric attitude error with the desired yaw set equal to the current yaw. 
By equating the desired yaw to the current yaw, we negate the dependency of yaw control, effectively creating a reduced attitude metric:
\begin{equation} \label{eq:e_1}
    \mathbf{e}_{cy} = \frac{1}{2} (\mathbf{R}_{des}^\top \mathbf{R} - \mathbf{R}^\top \mathbf{R}_{des})^{\vee} = \sin{\rho_e \mathbf{n}_e},
\end{equation}
where $\mathbf{R}_{des}$ is the desired attitude and $\mathbf{R}$ is the current estimated attitude. The terms $\rho_e$ and $\mathbf{n}_e$ are the axis angle representation of the error in attitude. 
With the current yaw metric, the error in rotation about the axis will go to zero within the system as the desired yaw is set to the current yaw. The \textit{vee map} $^\vee$ : $so(3) \rightarrow \mathbb{R}^{3}$ is the reverse hat map. 

\subsubsection{$\mathcal{S}^{2}$ Metric}
For the next error metric, we define the error in the tilt orientation using eq.~(\ref{eq:R_tilt}) as 
\begin{equation} \label{eq:R_tilt_er}
    \mathbf{e}_{s^2} = \frac{1}{2} (\mathbf{R}_{\phi_{des}}^\top \mathbf{R}_\phi - \mathbf{R}_\phi^\top \mathbf{R}_{\phi_{des}})^{\vee} = \sin{\rho_{\phi_e} \mathbf{n}_{\phi_e}},
\end{equation}
using only the tilt information of the given attitude of the rigid body.
This error metric is specifically designed for control around the $\mathbf{b}_1$ and $\mathbf{b}_2$ body axes, calculating errors exclusively in $\mathcal{S}^{2}$. 
This error metric in $\mathcal{S}^{2}$ aligns well to our problem at hand of only controlling attitude while surrendering control of the yaw angle.

\subsubsection{Thrust Vector Metric}
We set the unit vector that aligns with the quadrotor's thrust vector $\mathbf{n}_{tv}$ so that,
\begin{equation} \label{eq:n}
    \mathbf{n}_{tv} = \mathbf{R} \mathbf{e}_3.
\end{equation}
We can then define our last error metric as
\begin{equation} \label{eq:e_tv}
    \mathbf{e}_{tv} = \mathbf{R}^\top (\mathbf{n}_{{tv}_{des}} \times \mathbf{n}_{tv}).
\end{equation}
Here, $\mathbf{n}_{{tv}_{des}}$ signifies the unit vector along the anticipated thrust vector originating from the controller, while $\mathbf{n}$ represents the estimated thrust vector. 
This cross product dictates the necessary corrective direction required to align the real orientation with the intended orientation, making it an effective error metric.

\subsection{Fault Detection} \label{sec:fd}
We design, inspired by~\cite{mao2023propeller}, a fault detection method specifically using L1 adaptation~\cite{NairaL1} for rotor or propeller faults leading to a loss of effectiveness of the system. 
\begin{figure}
    \centering
    \includegraphics[width=0.9\linewidth, trim=0 800 0 900]{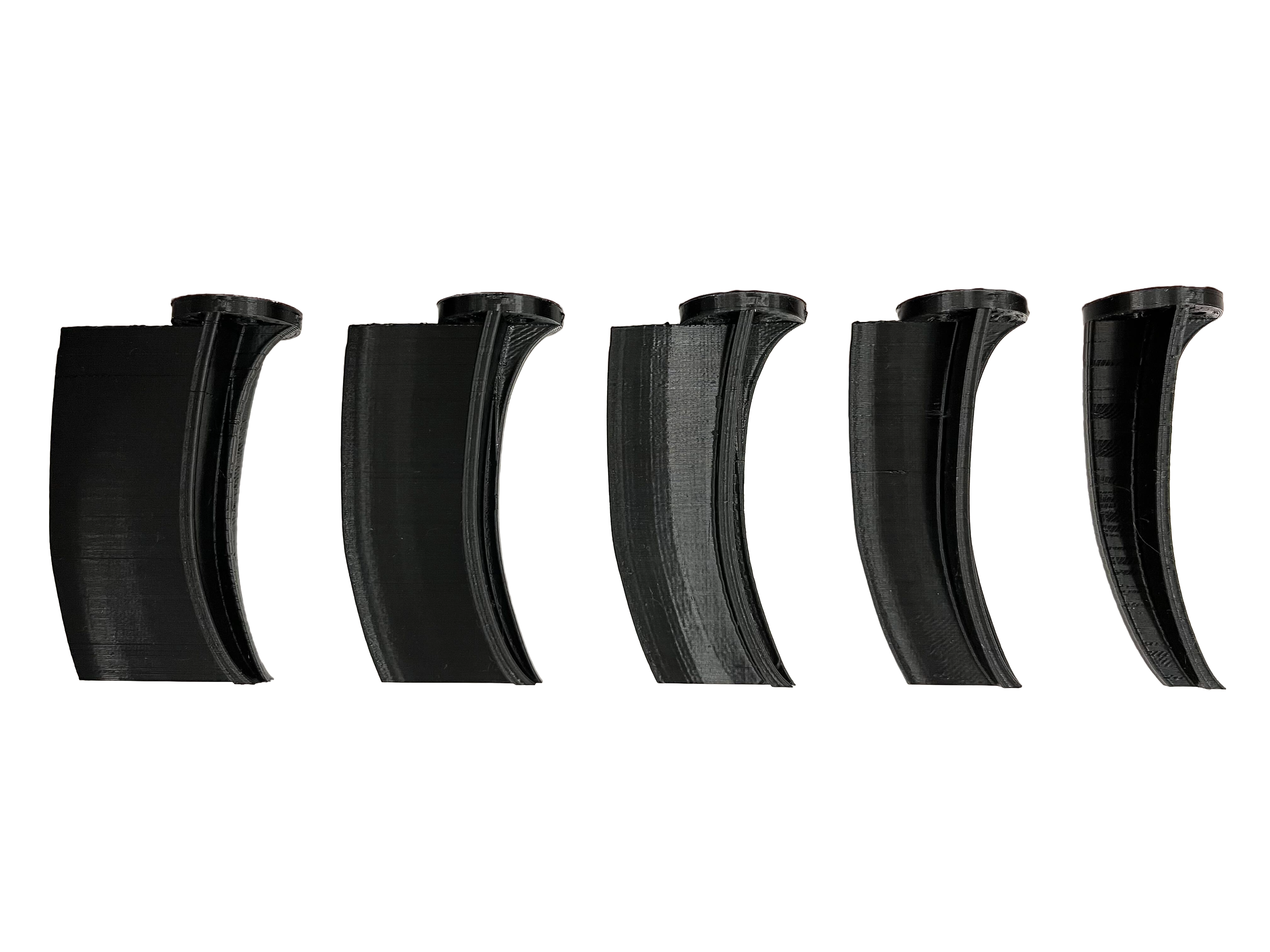}
    \caption{3D printed landing gears with added surface area to introduce rotational drag to the system.}
    \label{fig:landing_gear}
    \vspace{-0.3cm}
\end{figure}
First, the L1 augmented actions are calculated using the estimated external disturbances. The augmented actions are turned into RPMs. 
Then we use the ratio between the expected known RPM of each rotor-propeller pair calculated by our standard geometric controller and the L1 augmented RPM.
This relationship is used to initially guess the damage percentage of each of the rotor-propeller pair of the quadrotor, both identifying and estimating the amount of damage by
\begin{equation}
    d_i = 100
   \left(
    k_{f} \cdot \frac{\omega_{i}^2}{\omega_{L1i}^2}
    \right),
    \label{eq:damage_estimate}
\end{equation}
where $d_i$ is the estimated damage of the $i^\text{th}$ rotor-propeller pair, $k_f$ is the thrust coefficient, $\omega^2_i$ and $\omega_{L1i}^2$ are the standard and L1 augmented RPMs of the $i^\text{th}$ rotor-propeller pair.
We implement an inference methodology using quadratic programming using the initial estimate of propeller damage calculated in eq.~(\ref{eq:damage_estimate}) as a prior.
When a rotor-propeller pair is estimated to have damage over $50\%$, the controller transitions to the fault-tolerant controller.
The details of our fault detection algorithm can be found in~\cite{mao2023propeller}.

\subsection{Mechanical Design for Robust Fault-Tolerant Control}
In order for a quadrotor to produce sufficient thrust with $1$ or $2$ fully failed rotors, it must have at least $1$-to-$2$ thrust-to-weight ratio. 
This condition arises from the requirement that the two healthy rotors on opposite sides should collectively provide sufficient thrust to sustain the quadrotor's flight. 
Insufficient thrust (less than $1$-to-$2$ thrust-to-weight ratio) would render a fully fault-tolerant control scheme unattainable due to the absence of sufficient thrust, leading to a descent in altitude during hover.
\begin{figure}
    \centering
    \includegraphics[width=0.9\linewidth]{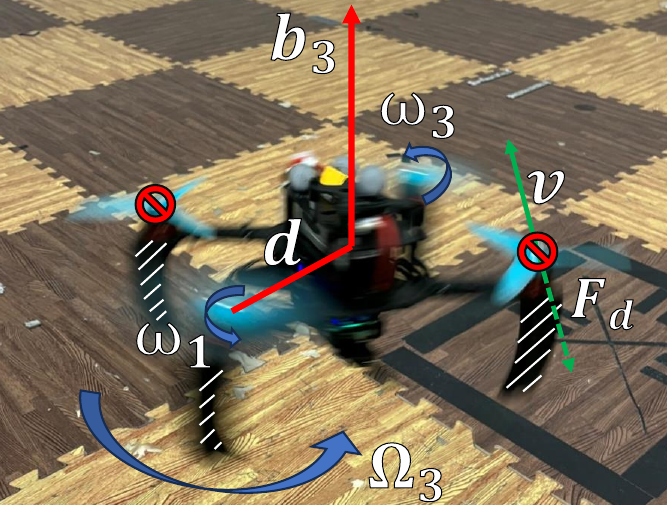}
    \caption{Quadrotor in fault-tolerant mode with variables related to drag coefficient estimation. The added surface area on the drone are highlighted with white hash marks.}
    \label{fig:drag_coef_picture}
    \vspace{-0.3cm}
\end{figure}
As our fault-tolerant controller surrenders the control of yaw, the faulty quadrotor will spin freely in yaw. 
Every quadrotor will have a steady yaw rate depending on its physical traits such as inertia and rpm of the healthy rotors. 
The highest yaw rate a system can handle will depend on the limit of the gyroscope used for state estimation. 
For instance, the gyroscope of our platform is rated up to $35$~rad/sec ($\pm 2000^\circ/\text{s}$) and will not be able to keep up if the quadrotor spins faster than the limit. 
To ensure we stay within the sensing limits of the gyro, we install landing gears with added surface area to slow the quadrotor's steady state yaw rate. 
Some examples of our design are shown in Fig.~\ref{fig:landing_gear}.
\begin{figure*}
    \centering
    \includegraphics[width=1\linewidth]{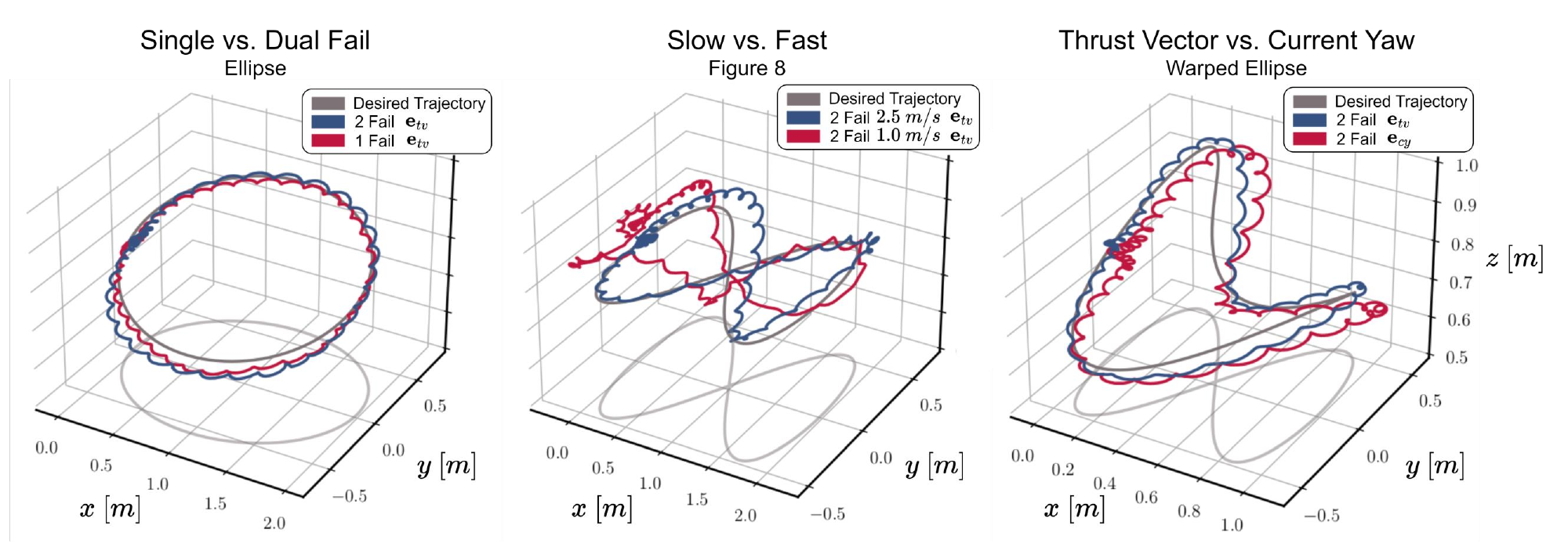}
    \caption{Showcase of three different trajectories and their tracking performance with different system attributes such as single vs. dual rotor failure cases, slow vs. fast trajectories, and different error metrics.}
    \label{fig:tracking}
    \vspace{-0.3cm}
\end{figure*}

While the increased surface area enhances the controllability of the fault-tolerant quadrotor, the results are tailored to our platform's inertia, geometric shape, and weight distribution. 
To offer findings applicable to a broader range of quadrotor platforms, we estimate the resulting rotational drag coefficient specifically for rotation about the $\mathbf{b}_3$ axis for the various rotational drag configurations.
All variables are depicted in Fig.~\ref{fig:drag_coef_picture} for clarity.
First we make the assumption that the yaw torque and drag about the $\mathbf{b}_3$ axis, introduced as $\tau_{\text{drag}}$ in eq.~(\ref{EoM}) are equivalent for the spinning quadrotor when it reaches a steady state yaw rate. 
Next we compute the linear velocity of the outer most point (at the rotor) of the quadrotor~$v$ by
\begin{equation}
    v = \Omega_3 d,
\end{equation}
where $d$ is the distance from the center of the mass of the quadrotor to the center of the rotor.
The drag force experienced by the same point is calculated by
\begin{equation}
    F_{\text{drag}} = \frac{\tau_{\text{drag}}}{4d},
\end{equation}
where $F_{\text{drag}}$ is the drag force.
Finally we estimate the drag coefficient $k_z$ about the $\mathbf{b}_3$ axis of the current configuration
\begin{equation}
\label{eq:drag}
    k_z = \frac{2 F_{\text{drag}}}{\rho A v^2} = \frac{2 F_{\text{drag}}}{\rho A (\Omega_3 d)^2},
\end{equation}
where $\rho$ is the air density coefficient and $A$ is the effective cross-sectional area of the quadrotor which can be determined by calculating the cross-section projection of the quadrotor in SolidWorks.
As the drag coefficient directly relates to the resulting yaw rate of the fault-tolerant quadrotor, we can use the drag coefficient as a design parameter.

\section{Results} \label{sec:results}
The experiments are conducted in a flying arena of $10\times6\times4~\si{m^3}$ at the Agile Robotics and Perception Lab (ARPL) at New York University.
The environment is equipped with a Vicon motion capture system that provides pose estimates at $100~\si{Hz}$. These estimates are fused with IMU measurements ($800~\si{Hz}$) through an Unscented Kalman Filter to provide state estimates at $500~\si{Hz}$.
The quadrotor, based on our previous work~\cite{LoiannoRAL2017}, is equipped with a VOXL\textsuperscript{\textregistered}2 ModalAI\textsuperscript{TM} specifically modified for a $2.5$ to $1$ thrust-to-weight ratio with a total weight of $700~\si{g}$.

With these experiments, we unveil the full potential of our active fault-tolerant controller, offering valuable insights into its capabilities and performance by answering the following questions: 
\begin{itemize}
    \item What are the dynamic limits of our fault-tolerant controller? How fast and aggressive can we fly?
    \item Which attitude error metrics perform best for fault-tolerant control?
    \item What are the reaction time and altitude required for successful and safe transition?
    \item What are the controllable bounds of the steady state rotational speed of the fault-tolerant quadrotor?
\end{itemize}
The investigation delves into these critical aspects of an active fault-tolerant control scheme, enhancing our understanding of the system and showcasing its capabilities in real-world scenarios.
\begin{table}[t!]
\centering
\caption {$xy$ plane tracking RMSE (in meters) for different trajectory speeds for single and dual rotor fails with three different error metrics.}
\begin{tabular}{cccc|ccc}
\toprule
\toprule
$\bm{v}_{\max}$ & \multicolumn{3}{c}{Single Fail} & \multicolumn{3}{c}{Dual Fail} \\
$[\SI{}{\meter/\second}]$ & $\mathbf{e}_{cy}$ & $\mathbf{e}_{S^2}$& $\mathbf{e}_{tv}$ & $\mathbf{e}_{cy}$ & $\mathbf{e}_{S^2}$& $\mathbf{e}_{tv}$\\
\midrule\midrule
0.5 & 0.041 & 0.040 & 0.038 & 0.057 & 0.066 & 0.061\\
1.0 & 0.077 & 0.078 & 0.087 & 0.089 & 0.099 & 0.094\\
1.5 & 0.135 & 0.137 & 0.101 & 0.139 & 0.145 & 0.148 \\
2.0 &  0.318 & 0.286  &  0.224 & 0.388  &  0.331 &  0.299 \\
2.5 & 0.338 & 0.312 &  0.309 & 0.329 & 0.316 & 0.356 \\
3.0 &  0.553 &  0.508 &  0.526 & 0.559  & 0.517  & 0.541\\
\bottomrule\bottomrule
\end{tabular}
\label{tab:xyrmse}
\vspace{-0.3cm}
\end{table}

\begin{figure*}
    \centering
    \includegraphics[width=0.95\linewidth]{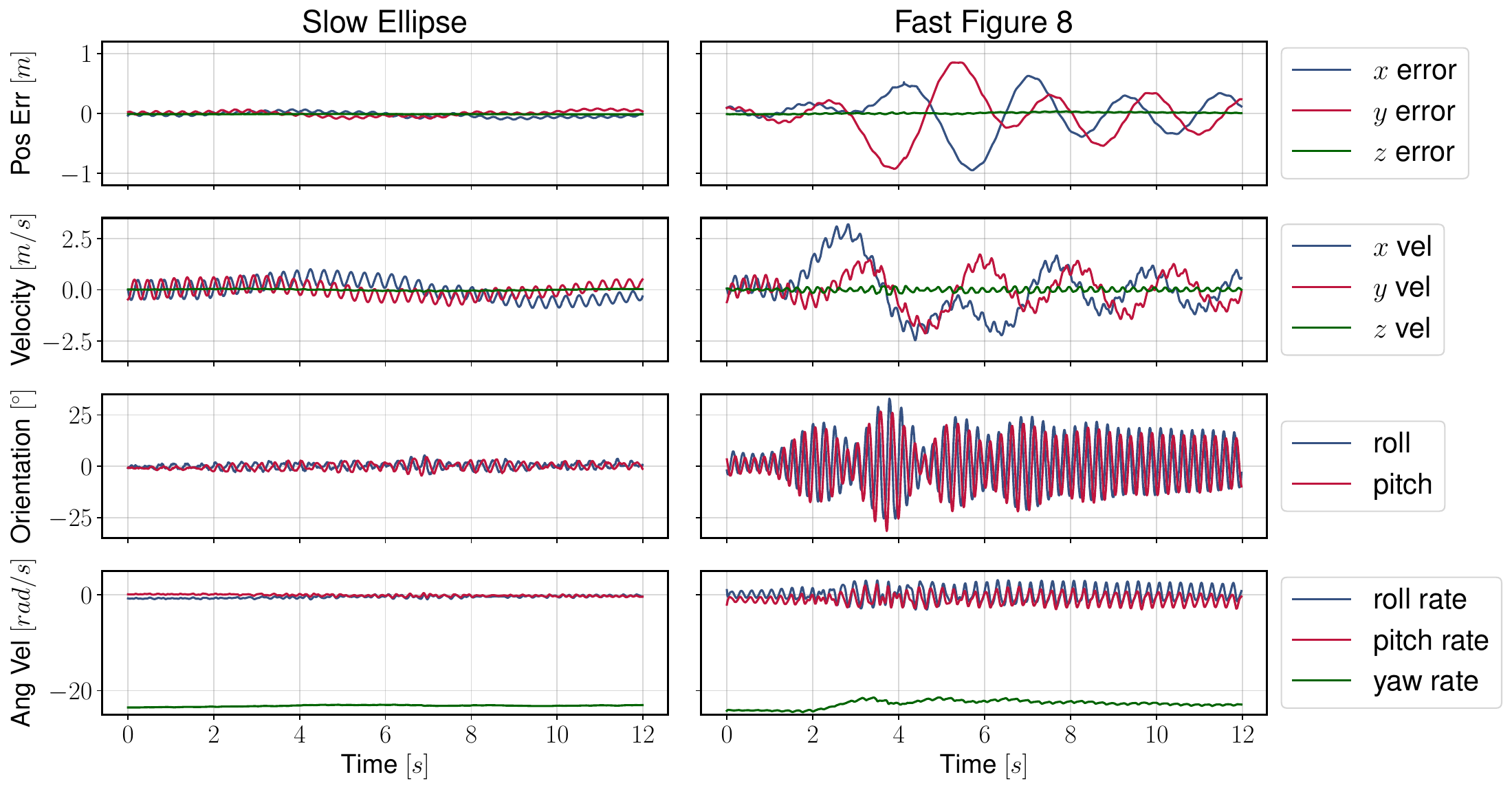}
    \caption{System dynamics of a fault-tolerant controlled quadrotor shown at two different speeds of an ellipse and figure 8 trajectory. Both are the single rotor failure case. Left column shows a slow and benign trajectory whereas the right column shows a fast and highly dynamic trajectory.}
    \label{fig:resulting_dynamics}
    \vspace{-0.3cm}
\end{figure*}

\subsection{Trajectory Tracking Performance}
We subject each of the error metrics to various types of trajectories to understand the full maneuverability of our controller with different error metrics. We test various types of trajectories including straight line, circle, figure 8, parabola ($xy$ and $xz$ planes), and warped ellipse ($xyz$ plane). 
We test each type of trajectory completed considering various speed profiles.
We report the tracking RMSE in Table~\ref{tab:xyrmse} categorized by test cases (single or dual rotor fail and error metric) and speed of the trajectory. 
We choose to show the $xy$ plane tracking RMSE since the $z$ tracking errors in all cases are similar.

From the results, we observe that the trajectory tracking error grows as the commanded linear velocity of the trajectory increases. 
The single rotor fail cases perform better than the dual rotor fail cases, corroborating the findings from simulations in~\cite{yeom2023geometric}.
When commanded speeds are slower, the tracking performance of our fault-tolerant controller is similar to the performance of a standard quadrotor with four healthy rotor-propeller pairs. 
Additionally, we notice a slightly higher position error for faster trajectories ($> 1.5~\si{m/s}$) with the current yaw error metric compared to the $\mathcal{S}^{2}$ and thrust vector error metrics. 
The current yaw metric may be the easiest to implement on an existing quadrotor controller, but tends to be less effective for aggressive trajectories.

A comparison considering different trajectories and design choices on the fault-tolerant controller are shown in Fig.~\ref{fig:tracking}.
The first plot shows the comparison of a single and dual failed rotor case with the thrust vector metric. 
The dual failed rotor case shows slightly higher position tracking errors.
The second plot shows the comparison of slow ($1.0~\si{m/s}$) and fast ($2.5~\si{m/s}$) tracking speeds for the dual failure case using the thrust vector error metric. 
The tracking error gets considerably worse with higher commanded velocities, barely resembling a figure 8.
Third plot shows the comparison of the thrust vector and current yaw metrics.
The current yaw error metric shows a slightly higher error when a trajectory requires quick and large changes in orientation.
Overall, the $\mathcal{S}^{2}$ and thrust vector methods perform similarly throughout the various tests.

The position error, velocity, roll and pitch angles as well as the angular velocities of our system for two different trajectories are shown in Fig.~\ref{fig:resulting_dynamics}.
The two trajectories are shown side by side to visually compare the differences in a slow and aggressive trajectory completed by the quadrotor. 
The left column shows a slow trajectory with very small position error and roll and pitch angles. 
General oscillations manifest in the overall dynamics even in a slow trajectory, where the quadrotor's pitch and roll moments translate into sinusoidal motions as the rigid body continuously spins at a steady rate. 
The right column shows an aggressive trajectory with heightened position errors as well as roll and pitch angles.
At its highest, the quadrotor rolls and pitches between $30 ^\circ$ and $-30 ^\circ$ rapidly within less than half a second, at a rate of approximately $3$ rad/sec, exhibiting an extremely dynamic maneuver.

We find the maximum speed of our fault-tolerant controller to be $3~\si{m/s}$ and the maximum attitude angle (roll, pitch) to be approximately $30^\circ$. 
The maximum speed is due to the current size of our indoor arena, and the system is capable of greater speeds with the room for further acceleration.
However, a flight with attitude angles greater than $30^\circ$ is ill-advised as it has potential for loss of control or crash as a spinning body with $30 ^\circ$ oscillation in an additional axis is already highly dynamic. 
Moreover, engaging in such dynamic maneuvers results in position errors as large as a full meter along one Cartesian axis.
For optimal safety and performance, we recommend adopting slow nominal flights when utilizing a fault-tolerant controller. 
This approach is particularly prudent in situations where there is no imperative need to follow highly dynamic trajectories, especially during the occurrence of a rotor failure.

\begin{figure}
    \centering
    \vspace{-0.3cm}
    \includegraphics[width=1\linewidth, trim=0 10 0 0, clip]{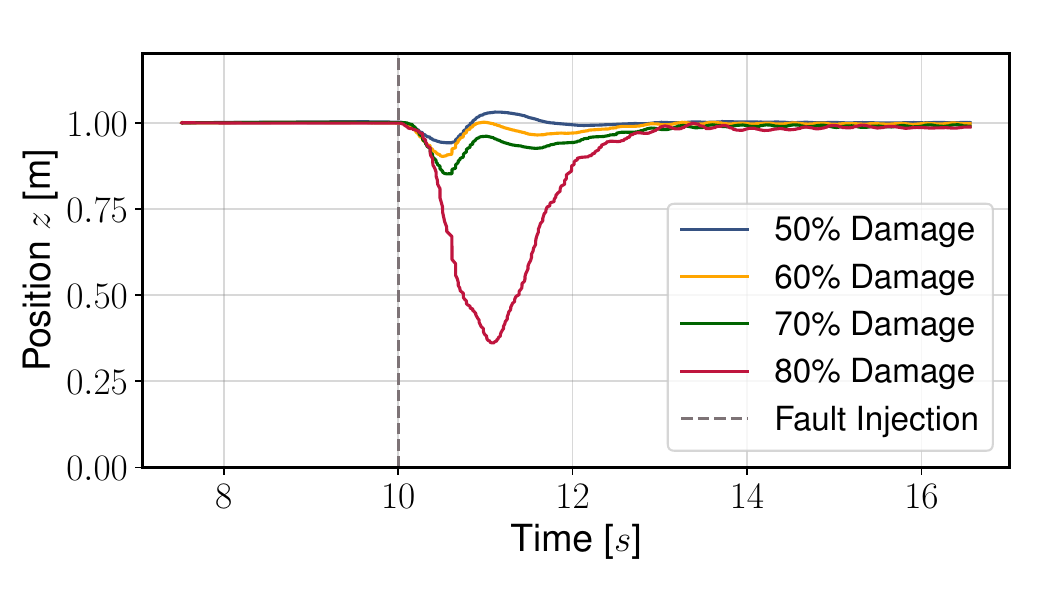}
    \caption{System altitude of a quadrotor during fault injection, detection, and controller transition for $50$, $60$, $70$, and $80\%$ damage injections on a single rotor.}
    \label{fig:transition}
    \vspace{-0.5cm}
\end{figure}
\subsection{Fault Detection}
We test the effectiveness of our active fault-tolerant controller by artificially injecting rotor damage, lowering the affected rotor's RPM by a set multiplier.
Fig.~\ref{fig:transition} shows the $z$ position, or altitude of the quadrotor as various degrees of rotor damage are injected and the controller transitions to the fault-tolerant version. 
For the $50$, $60$, and $70\%$ damage cases, the altitude losses were measured at $0.05~\si{m}$, $0.08~\si{m}$, and $0.13~\si{m}$, respectively. 
In contrast, the $80\%$ damage case exhibited a significantly larger altitude loss of $0.6 \si{m}$. 
This substantial decrease in altitude can be attributed to the impaired rotor's inability to generate significant thrust, even with the L1 adaptation mechanism in place.
Hence we discover the limit of L1 adaptation, wherein adaptation alone is unable to provide sufficient actuation for maintaining stable flight in the presence of a faulty rotor.
This is seen by the quadrotor falling dramatically as L1 adaptation is unable to boost the faulty rotor's thrust sufficiently to sustain flight.
Overall, the latency or reaction time from a fault injection to actual detection must be in the $100-150~\si{ms}$ range.

\begin{figure}
    \centering
    \vspace{-0.2cm}
    \includegraphics[width=1\linewidth, trim=0 10 0 0, clip]{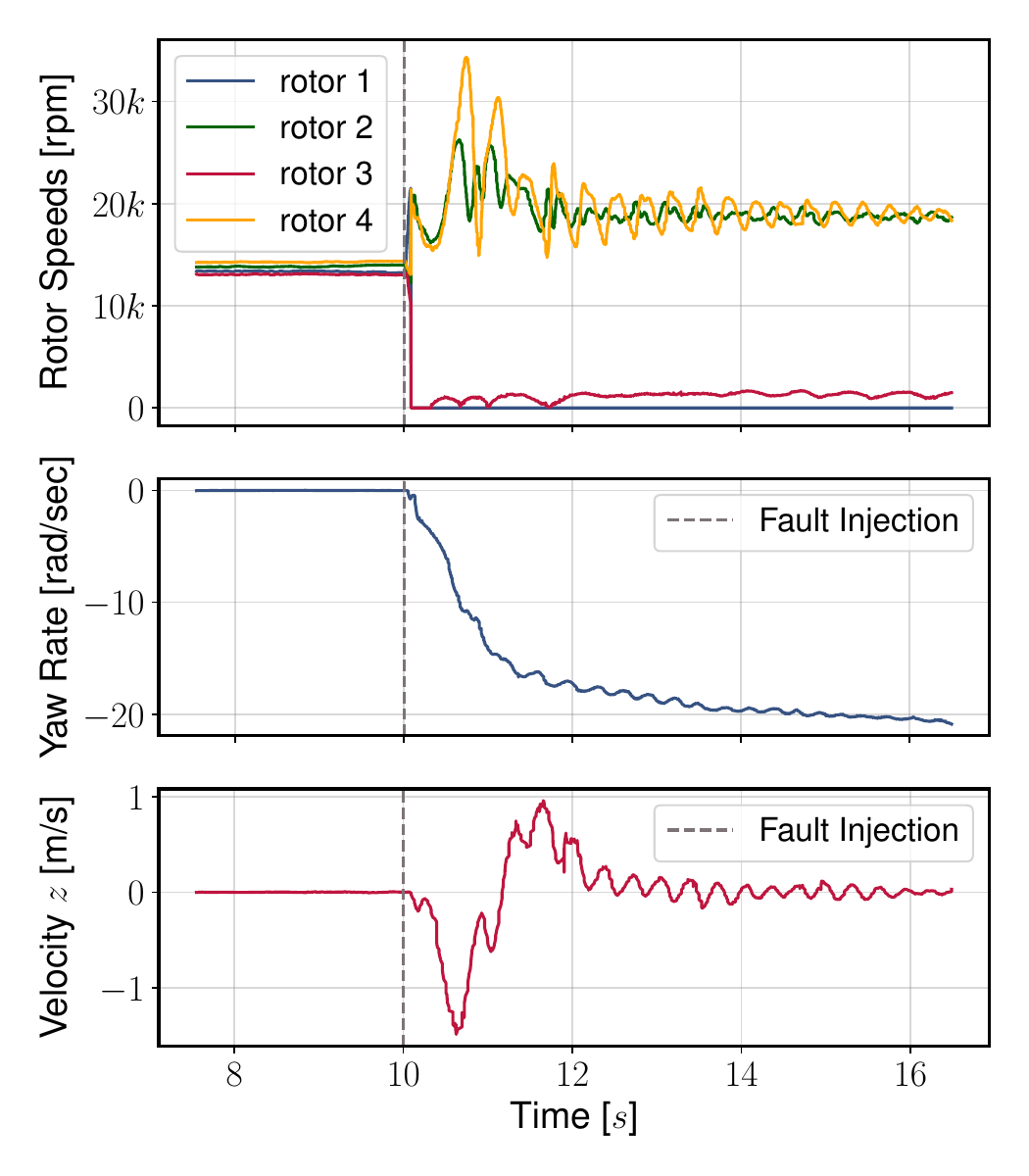}
    \caption{Commanded rotor speeds, yaw rate, and velocity in~$z$ of a quadrotor when $80\%$ damage is injected to rotor $\#1$.}
    \label{fig:rpms}
    \vspace{-0.2cm}
\end{figure}

Fig.~\ref{fig:rpms} shows the commanded rotor speeds, yaw rate build-up, and velocity in $z$ of the quadrotor when $80\%$ damage is injected to rotor-propeller pair $\#1$. 
At $80\%$ damage, the rotor is slowed down to idle RPM of less than $5000$ RPM, contributing essentially no thrust to the system.
The gray dashed line in both Fig.~\ref{fig:transition} and \ref{fig:rpms} indicates when the fault is injected at $10$ seconds.
The rotor speeds show the fault detection and reaction are almost immediate ($100~\si{ms}$) to the fault injection, as the speeds of rotors $2$ and $4$ spike up to approximately $20,000$ rpm.
While the fault detection time is rapid, the quadrotor descends due to insufficient thrust, we find that the controller requires significant spin-up time for the yaw rate to attain a controllable level and simultaneously correct its orientation to generate upward thrust.
A controllable yaw rate is found to be greater than $10$~rad/sec and is addressed in the next subsection of this paper.
Due to time necessary for the quadrotor to start spinning at controllable rate, data in Figs.~\ref{fig:transition} and \ref{fig:rpms} show the quadrotor falling for almost a full second before the quadrotor is able to recover to controlled flight. 

In order for successful transition, the quadrotor must have sufficient altitude.
As seen in our $80\%$ damage example, it is possible for a quadrotor to loose up to $0.6~\si{m}$ during a transition and recovery to fault-tolerant control.
Therefore the authors recommend users to fly their quadrotors at least $1~\si{m}$ above the ground.  
We found our fault detection latency to be similar to or better than the state of the art~\cite{schijndel2023eject,Avram_2017_detect_adapt, Madruga2023LOEAero}.
A faster detection would aid in system performance; however, what actually causes the loss in altitude in cases of full loss of thrust is the transition time to spin-up to a controllable yaw rate.
More powerful rotors may be able to get the quadrotor to its steady state quicker, resulting in smaller altitude loss in the transition to a fault-tolerant controller.

\begin{figure}
    \centering
    \includegraphics[width=0.99 \linewidth]{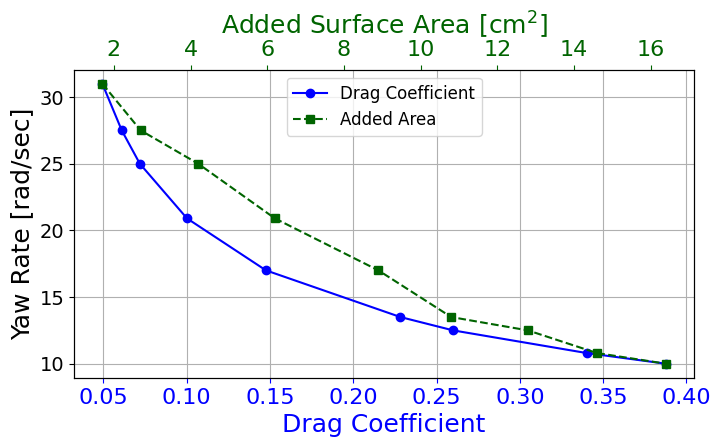}
    \caption{Parametric study of steady state yaw rate with respect to added surface area as well as the estimated drag coefficient for each case.}
    \vspace{-0.5cm}
    \label{fig:yawrate}
\end{figure}

\subsection{Mechanical Design Evaluation}
The rapid spin rate exhibited by the quadrotor upon activation of the fault-tolerant control is not suitable nor desirable for safety of flight. 
In fault-tolerant mode, the remaining healthy rotors achieve speeds exceeding $20,000$~rpm, leading to a yaw rate surpassing $25$~rad/sec for the spinning quadrotor.
Such behavior is not only concerning but also potentially hazardous if not effectively controlled. 
Considering the alarming nature of this quadrotor's behavior, it becomes imperative to explore potential solutions such as incorporating rotational drag into the quadrotor's design. 
We choose to add surface area to the landing gears, effectively adding to the system's rotational drag. 
In this way, the yaw rate during fault-tolerant control can be reduced, enhancing the overall safety of the system. 
In an effort to characterize the relationship between rotational drag and steady state yaw rate of our quadrotor, we incrementally add small amounts of rotational drag to our quadrotor by adding surface area to our landing gears. 
This is a simple and cheap solution to study the rotational dynamics of the system. 

The relationship between added surface area and yaw rate is presented in Fig.~\ref{fig:yawrate}.
This data was gathered experimentally by adding incremental amounts of surface area to the landing gears and turning on the fault-tolerant mode in flight, measuring the resulting steady state yaw rate.
The quadrotor is uncontrollable outside of the bounds of the graph. 
Below approximately $10~\si{rad/sec}$ yaw rate, the quadrotor becomes extremely sensitive to weight asymmetries; as our quadrotor is slightly back-heavy, the slower yaw rate caused large oscillations in roll and pitch eventually causing the quadrotor to crash. 
Even if the quadrotor is perfectly symmetrical, it would not have enough control authority to effectively orient itself to follow a trajectory when it is spinning at less than $10~\si{rad/sec}$. 
This is a testament to how the angular momentum and rotational inertia of a fault-tolerant quadrotor actually aids to its overall system stability.

To conduct a more comprehensive analysis, we estimate the drag coefficient of the stable region of augmented drag configurations with the method outlined in Section~\ref{sec:methodology}.D. 
The results of the relationship between the estimated drag coefficient and yaw rate are depicted in Fig.~\ref{fig:yawrate}. 
The observed relationship between the estimated drag coefficient and yaw rate is quadratic, which corroborates the analytical relationship as shown in eq.~(\ref{eq:drag}).
The results demonstrate that, to maintain a controllable steady-state yaw rate, the quadrotor's rotational drag coefficient should fall within the range of $0.05$ to $0.35$, irrespective of the platform.
These drag coefficients need to be respected to ensure a controllable fault-tolerant quadrotor while surrendering the control of yaw. 
This finding underscores the versatility of the drag coefficient as a platform-agnostic metric potentially applicable to any quadrotor.
This holds true for any quadrotor platform equipped with a gyroscope featuring a nominal rate limit of $\pm 2000^\circ/\text{s}$ or $35$ rad/sec, a standard specification for MEMS (Micro-Electro-Mechanical Systems) devices in quadrotors.
For example, in order for a fault-tolerant quadrotor to be within its gyro limit, it must have a less than $35$ rad/sec yaw rate. This means the system must have a drag coefficient of approximately $0.05$ or higher as seen in Fig.~\ref{fig:yawrate}.

\section{Conclusion} \label{sec:conclusion}
In this paper, we showed a design approach for quadrotor fault-tolerant control. It combines an effective fault detection system with a robust control strategy, complemented by a careful mechanical design to ensure the platform actuator and sensing constraints are respected. This integrated approach ensures the necessary level of robustness without any additional sensors or computation load. The proposed approach has the potential to be applied to any type of quadrotor.

Future works will focus on motion capture free fault-tolerant control solutions with additional fault detection methods along with control solutions for less dynamic transition periods. 

\bibliographystyle{IEEEtran}
\bibliography{references}

\begin{thebibliography}{10}
\providecommand{\url}[1]{#1}
\csname url@samestyle\endcsname
\providecommand{\newblock}{\relax}
\providecommand{\bibinfo}[2]{#2}
\providecommand{\BIBentrySTDinterwordspacing}{\spaceskip=0pt\relax}
\providecommand{\BIBentryALTinterwordstretchfactor}{4}
\providecommand{\BIBentryALTinterwordspacing}{\spaceskip=\fontdimen2\font plus
\BIBentryALTinterwordstretchfactor\fontdimen3\font minus
  \fontdimen4\font\relax}
\providecommand{\BIBforeignlanguage}[2]{{%
\expandafter\ifx\csname l@#1\endcsname\relax
\typeout{** WARNING: IEEEtran.bst: No hyphenation pattern has been}%
\typeout{** loaded for the language `#1'. Using the pattern for}%
\typeout{** the default language instead.}%
\else
\language=\csname l@#1\endcsname
\fi
#2}}
\providecommand{\BIBdecl}{\relax}
\BIBdecl

\bibitem{almurib2011searchandrescue}
H.~A.~F. Almurib, P.~T. Nathan, and T.~N. Kumar, ``Control and path planning of
  quadrotor aerial vehicles for search and rescue,'' in \emph{SICE Annual
  Conference}, 2011, pp. 700--705.

\bibitem{li2021transportation}
G.~Li, R.~Ge, and G.~Loianno, ``Cooperative transportation of cable suspended
  payloads with mavs using monocular vision and inertial sensing,'' \emph{IEEE
  Robotics and Automation Letters}, vol.~6, no.~3, pp. 5316--5323, 2021.

\bibitem{yeom2023geometric}
J.~Yeom, G.~Li, and G.~Loianno, ``Geometric fault-tolerant control of
  quadrotors in case of rotor failures: An attitude based comparative study,''
  \emph{IEEE/RSJ International Conference on Intelligent Robots and Systems
  (IROS)}, 2023.

\bibitem{mao2023propeller}
J.~Mao, J.~Yeom, S.~Nair, and G.~Loianno, ``From propeller damage estimation
  and adaptation to fault tolerant control: Enhancing quadrotor resilience,''
  \emph{arXiv preprint arXiv:2310.13091}, 2023.

\bibitem{abbaspour2020survey}
A.~Abbaspour, S.~Mokhtari, A.~Sargolzaei, and K.~K. Yen, ``A survey on active
  fault-tolerant control systems,'' \emph{Electronics}, vol.~9, no.~9, p. 1513,
  2020.

\bibitem{chaturvedi2011attitudecontrol}
N.~A. Chaturvedi, A.~K. Sanyal, and N.~H. McClamroch, ``Rigid-body attitude
  control,'' \emph{IEEE Control Systems Magazine}, vol.~31, no.~3, pp. 30--51,
  2011.

\bibitem{lippiello2014emergency}
V.~Lippiello, F.~Ruggiero, and D.~Serra, ``Emergency landing for a quadrotor in
  case of a propeller failure: A pid based approach,'' in \emph{IEEE
  International Symposium on Safety, Security, and Rescue Robotics (SSRR)},
  2014, pp. 1--7.

\bibitem{freddi2011feedback}
A.~Freddi, A.~Lanzon, and S.~Longhi, ``A feedback linearization approach to
  fault tolerance in quadrotor vehicles,'' \emph{IFAC proceedings volumes},
  vol.~44, no.~1, pp. 5413--5418, 2011.

\bibitem{sharifi2010slidingmode}
F.~Sharifi, M.~Mirzaei, B.~W. Gordon, and Y.~Zhang, ``Fault tolerant control of
  a quadrotor uav using sliding mode control,'' in \emph{Conference on Control
  and Fault-Tolerant Systems (SysTol)}, 2010, pp. 239--244.

\bibitem{deCrousaz2015sql}
C.~de~Crousaz, F.~Farshidian, M.~Neunert, and J.~Buchli, ``Unified motion
  control for dynamic quadrotor maneuvers demonstrated on slung load and rotor
  failure tasks,'' in \emph{IEEE International Conference on Robotics and
  Automation (ICRA)}, 2015, pp. 2223--2229.

\bibitem{mueller2014lqr}
M.~W. Mueller and R.~D'Andrea, ``Stability and control of a quadrocopter
  despite the complete loss of one, two, or three propellers,'' in \emph{IEEE
  International Conference on Robotics and Automation (ICRA)}, 2014, pp.
  45--52.

\bibitem{sun2021indi}
S.~Sun, X.~Wang, Q.~Chu, and C.~d. Visser, ``Incremental nonlinear
  fault-tolerant control of a quadrotor with complete loss of two opposing
  rotors,'' \emph{IEEE Transactions on Robotics}, vol.~37, no.~1, pp. 116--130,
  2021.

\bibitem{nan2022npmcfault}
F.~Nan, S.~Sun, P.~Foehn, and D.~Scaramuzza, ``Nonlinear mpc for quadrotor
  fault-tolerant control,'' \emph{IEEE Robotics and Automation Letters},
  vol.~7, no.~2, pp. 5047--5054, 2022.

\bibitem{Madruga2023LOEAero}
S.~P. Madruga, T.~P. Nascimento, F.~Holzapfel, and A.~M.~N. Lima, ``Estimating
  the loss of effectiveness of uav actuators in the presence of aerodynamic
  effects,'' \emph{IEEE Robotics and Automation Letters}, vol.~8, no.~3, pp.
  1335--1342, 2023.

\bibitem{Avram_2017_detect_adapt}
R.~C. Avram, X.~Zhang, and J.~Muse, ``Quadrotor actuator fault diagnosis and
  accommodation using nonlinear adaptive estimators,'' \emph{IEEE Transactions
  on Control Systems Technology}, vol.~25, no.~6, pp. 2219--2226, 2017.

\bibitem{schijndel2023eject}
B.~A.~S. van Schijndel, S.~Sun, and C.~C. de~Visser, ``Fast loss of
  effectiveness detection on a quadrotor using onboard sensors and a kalman
  estimation approach,'' in \emph{International Conference on Unmanned Aircraft
  Systems (ICUAS)}, 2023, pp. 1--8.

\bibitem{vibrations_diagonosis}
B.~Ghalamchi, Z.~Jia, and M.~W. Mueller, ``Real-time vibration-based propeller
  fault diagnosis for multicopters,'' \emph{IEEE/ASME Transactions on
  Mechatronics}, vol.~25, no.~1, pp. 395--405, 2020.

\bibitem{luLOE}
P.~Lu and E.-J. van Kampen, ``Active fault-tolerant control for quadrotors
  subjected to a complete rotor failure,'' in \emph{IEEE/RSJ International
  Conference on Intelligent Robots and Systems (IROS)}, 2015, pp. 4698--4703.

\bibitem{lee2010so3}
T.~Lee, M.~Leok, and N.~H. McClamroch, ``Geometric tracking control of a
  quadrotor uav on se(3),'' in \emph{49th IEEE Conference on Decision and
  Control (CDC)}, 2010, pp. 5420--5425.

\bibitem{NairaL1}
Z.~Wu, S.~Cheng, K.~A. Ackerman, A.~Gahlawat, A.~Lakshmanan, P.~Zhao, and
  N.~Hovakimyan, ``L1adaptive augmentation for geometric tracking control of
  quadrotors,'' in \emph{2022 International Conference on Robotics and
  Automation (ICRA)}, 2022, pp. 1329--1336.

\bibitem{LoiannoRAL2017}
G.~Loianno, C.~Brunner, G.~McGrath, and V.~Kumar, ``Estimation, control, and
  planning for aggressive flight with a small quadrotor with a single camera
  and imu,'' \emph{IEEE Robotics and Automation Letters}, vol.~2, no.~2, pp.
  404--411, April 2017.

\end{thebibliography}
\end{document}